\documentclass{article}

\usepackage[english]{babel}
\usepackage{amsthm}
\usepackage{float}
\usepackage{amsmath} 
\usepackage{mathtools}
\usepackage{amssymb}
\usepackage{xfrac} 
\usepackage[utf8]{inputenc} 
\usepackage[T1]{fontenc}    
\usepackage{hyperref}       
\usepackage{url}            
\usepackage{booktabs}       
\usepackage{amsfonts}       
\usepackage{nicefrac}       
\usepackage{microtype}      
\usepackage{lipsum}
\usepackage{fancyhdr}       
\usepackage{graphicx}       
\usepackage{xcolor}         
\usepackage{subcaption}
\usepackage{booktabs}

\usepackage{todonotes}

\newcommand{\dd}{\mathrm d}

\usepackage[preprint]{neurips_2026}

\usepackage[capitalize,noabbrev]{cleveref}

\theoremstyle{plain}
\newtheorem{theorem}{Theorem}[section]

\newtheorem{lemma}[theorem]{Lemma}

\theoremstyle{definition}
\newtheorem{definition}[theorem]{Definition}

\theoremstyle{remark}
\newtheorem{remark}[theorem]{Remark}

\newcommand{\prefactor}[2]{A_{#1,#2}}
\DeclareMathOperator*{\argmin}{arg\,min}

\title{Global Optimization via Softmin Energy Minimization}

\author{
Samuele Saviozzi \\
Department of Mathematics\\
University of Pisa\\
Pisa, Italy \\
\texttt{samuele.saviozzi@phd.unipi.it} \\
\And
Andrea Agazzi \\
Institute of Mathematical Statistics \\and Actuarial Science\\ 
University of Bern\\
Bern, Switzerland\\
\texttt{andrea.agazzi@unibe.ch} \\
\AND
Vittorio Carlei \\
Department of Buisness Economics\\ 
University Gabreiele D'annunzio\\
Chieti, Italy
\texttt{carlei@unich.it} \\
\And
Marco Romito \\
Department of Mathematics\\
University of Pisa\\
Pisa, Italy \\
\texttt{marco.romito@unipi.it} \\
}

\begin{document}











\maketitle

\begin{abstract}
    Non-convex optimization in high dimensions requires balancing gradient information with global exploration, a tension that Langevin dynamics and other gradient-informed methods address only partially. We introduce a gradient-based swarm method driven by a Softmin Energy interaction $J_\beta(\mathbf{x})$, a smooth approximation of the per-particle minimum that couples particles through their relative energies. Combining the resulting gradient flow with Brownian noise and an annealing schedule on $\beta$, we obtain a dynamics that retains the efficiency of gradient steps while inheriting the global-search behavior of swarm methods. Our main theoretical result is an Eyring-Kramers-type estimate showing that the effective potential barriers are strictly smaller than those of overdamped Langevin dynamics, yielding provably faster transitions between local minima. Numerical experiments on double-well and high-dimensional Ackley benchmarks confirm the analysis and show improved convergence over Simulated Annealing, Gradient-Enhanced Particle Swarm Optimization, and Gradient-Aware Consensus Based Optimization.
\end{abstract}

\section{Introduction}
Non-convex optimization problems are notoriously difficult, as the presence of multiple local minima can trap gradient-based algorithms. Existing approaches such as gradient descent perform well for convex functions with a single minimum but will likely converge to suboptimal local minima in more complex landscapes. This challenge is especially prominent in machine learning, where training deep neural networks on non-convex losses often leads to inconsistent performance across random initializations. Similarly, in reinforcement learning, escaping sharp local minima remains a key bottleneck for achieving state-of-the-art results. 

A classical strategy to address this issue is to add a noise term to the dynamics, allowing the system to escape local minima and leading to stochastic gradient–based methods, for example Langevin dynamics \cite{CoffeyKalmykovWaldron2004}. Stochastic gradient-based methods for optimization, while effective, typically exhibit poor transition times between local minima. The escape rate of a particle subject to Gaussian noise from a local minimum depends exponentially on the height of the potential barrier surrounding that minimum \cite{berglund2011kramers}. Methods that improve convergence rates over gradient-based algorithms often rely on additional assumptions about the objective function $f$. For example, if $f$ admits a decomposition as a linear combination of simpler functions $\psi_i$, i.e., $f(x) = \sum_i \psi_i(x)$, then the Method of Moments \cite{meziat2003method} can be employed to converge to the global minimum. This type of decomposition is particularly useful in areas such as signal processing and machine learning, where complex functions are represented as sums of basis functions, such as Fourier series or wavelets \cite{shao1993signal}. In more general settings where such structure is not available, Simulated Annealing and its variants \cite{bolte2024swarm,xu2018global} are widely used, leveraging a decaying temperature schedule to shift the dynamics from noisy exploration to refined local search.

A natural extension of stochastic gradient-based approaches is the use of multiple particles to explore the search space in parallel. For instance, when particles evolve independently under Langevin dynamics, the expected transition times between local minima decrease compared to single-particle dynamics purely due to statistical averaging over multiple independent trajectories. Moreover, allowing particles to cooperate rather than evolve independently can further improve exploration and convergence. The idea of particle cooperation is often central in Metaheuristic methods \cite{bandaru2016metaheuristic,yang2020nature}. Such methods provide a broad framework for global optimization and have demonstrated strong empirical performance, although establishing rigorous theoretical convergence guarantees is often challenging. Many such methods employ a particle-swarm-like paradigm \cite{hussain2019metaheuristic,rajwar2023exhaustive}, in which multiple interacting particles explore the search space in parallel to locate global minimizers.

The global optimization method proposed in this paper builds on these ideas by combining stochastic gradient techniques with cooperative multi-particle dynamics. We employ the well-known Softmin Energy functional $J_\beta$, which smoothly interpolates between the average and the hard minimum of the function values across particle positions. This functional acts as the source of interaction among particles, whose evolution is governed by its gradient, driving the system toward more effective exploration and optimization.

The Softmin function provides a natural way to interpolate between uniform weighting and hard minima via temperature $\beta>0$. A common use of the Softmin function is to define the Gibbs distribution, where the probability of sampling $x_i$ is given by: 
\[
    \mathbb{P}[i] = \tfrac{\exp{(-\beta{x_i})}}{\sum_{j=1}^n \exp{(-\beta x_j})},
\] 
with $\beta\to\infty$ concentrating on $\min_{i} x_i$ and $\beta\to0$ uniform \cite{bertsekas1997nonlinear}. This underlies stochastic optimization \cite{ackley1985boltzmann, srivastava2012multimodal, gelfand1990sampling,gelfand2000gibbs} and exploration in Reinforcement Learning \cite{pan2019reinforcement, haarnoja2017reinforcement}.

Our proposed method is a stochastic gradient-based particle swarm approach that leverages a Softmin-type energy functional to effectively incorporate gradient information from the particles. Consider a differentiable function $f:\mathbb{R}^d\to\mathbb{R}$. The objective is to approximately solve the global minimization problem $\min_{x\in\mathbb{R}^d}f(x)$. Given a population of $n$ particles $\mathbf{x}=(x_1,\dots,x_n)$ with $x_i\in\mathbb{R}^d$ for all $i\in[n]$, the Softmin Energy is defined as: 
\begin{equation}
    J_\beta(\mathbf{x})=\sum_{i=1}^n\frac{\exp(-\beta f(x_i))}{\sum_{j=1}^n\exp(-\beta f(x_j))}f(x_i),
    \label{eq:Softmin-energy}
\end{equation}
which provides a smooth approximation of the empirical minimum among the $f(x_i)$. The parameter $\beta>0$ controls the approximation's sharpness: larger values yield a closer approximation to the hard minimum $\min_i f(x_i)$, while $\beta\to0$ recovers the average $\frac 1 n \sum_{i=1}^n f(x_i)$. 

To minimize $J_\beta$ we consider the stochastic gradient flow in the particle space $\mathbb{R}^{nd}$: 
\begin{equation}\label{e:sde}
   d\mathbf{x}_t = -n\nabla J_{\beta_t}(\mathbf{x}_t)dt+\sqrt{2\sigma}dB_t, 
\end{equation} 
where $(B_t)_{t\geq 0}$ represents a standard $nd$-dimensional Brownian motion and $\sigma>0$ introduces stochasticity to promote exploration and escape from local minima.

\subsection{Contribution} 
The main novelty of this work is the rigorous analysis of stochastic gradient-based dynamics driven by the Softmin Energy functional, $J_\beta(\mathbf{x})$. In this paper, we prove that the resulting inter-particle interaction mechanism provides strict theoretical advantages over standard gradient-based and non-interacting methods for non-convex global optimization.

\paragraph{Dynamic Ensemble Partitioning and Accelerated Convergence} The interaction induced by $\nabla J_\beta$ naturally partitions the particle ensemble via an effective learning rate, $\prefactor{t}{k}$. Particles below the Softmin threshold ($\prefactor{t}{k} > 0$) converge to local minima, while trapped, high-energy particles ($\prefactor{t}{k} < 0$) are automatically repelled to hill-climb and explore (Theorem \ref{thm:stationary-points}). Furthermore, we prove that trajectories in the strongly minimizing regime ($\prefactor{t}{k} \ge 1$) are invariant and yield exponential convergence rates of $\mathcal{O}(e^{-\lambda t})$ under local $\lambda$-strong convexity (Theorem \ref{thm:rate-of-convergence-minimizing-particle}), strictly outperforming or matching standard overdamped Langevin dynamics.
    
\paragraph{Provable Topological Barrier Flattening} We demonstrate that the Softmin dynamics do not merely reduce effective potential barriers, but analytically cap them. By integrating the effective drift, we prove that arbitrarily high physical barriers escaping a local minimum are flattened to a constant maximum of $\mathcal{O}(1/\beta)$ (Theorem \ref{thm:transition-time}). Under the strict inverse temperature condition $\beta < (2+\sqrt{3})/f^*$, this global barrier-flattening guarantees exponentially faster expected transition times between basins in the small-noise regime governed by Kramers' law. This structural landscape modulation is a direct consequence of the continuous Softmin repulsion, distinguishing our approach from generic stochastic swarm methods and Simulated Annealing.

\subsection{Related work}

\paragraph{Metaheuristic optimization methods} 
Metaheuristic optimization methods include Consensus-Based Optimization (CBO) \cite{CarrilloChoiTotzeckTse2018,fornasier2024consensus,pinnau2017consensus, carrillo2021consensus}, Genetic Algorithms \cite{Goldberg1989}, Ant Colony Optimization \cite{DorigoStutzle2004}, and Particle Swarm Optimization (PSO) \cite{kennedy1995particle, noel2012new}. Traditional particle-swarm approaches typically operate without explicit gradient information, relying instead on position-based updates to coordinate particles toward promising regions. When reliable gradient information is available, leveraging it enables convergence rates comparable to standard stochastic gradient descent, which gradient-agnostic metaheuristics cannot achieve.

Recent gradient-based swarm methods \cite{riedl2023gradient, bolte2024swarm, tang2024discrete, carrillo2021consensus,noel2012new}, have made progress by incorporating gradient information into multi-particle dynamics. However, our approach differs fundamentally in its mechanism for inducing exploration and escaping barriers. Rather than consensus-based averaging or general gradient-weighted potentials, we employ the Softmin energy functional to explicitly create a minimizing/maximizing partition of the particle ensemble: particles below the Softmin threshold are drawn toward local minima via gradient descent, while particles above the threshold are repelled toward saddles and maxima. This automatic role assignment (Theorem \ref{thm:stationary-points}) simultaneously reduces effective potential barriers (Theorem \ref{thm:transition-time}) and forces suboptimal particles into exploratory regimes without requiring manual basin-switching heuristics.

To the best of our knowledge, no other metaheuristic employs the Softmin energy functional, computed as the softmin of function values at particle positions, as the sole source of inter-particle interaction. This design enables provable faster escape times from local maxima and provable linear convergence rates to minima in the small-noise limit, as demonstrated in Section \ref{sec:hitting-times} and validated experimentally on benchmark landscapes where barriers dominate the hardness.

\paragraph{Connection to Statistical Physics and Functional Design}
Our approach shares conceptual roots with population-based statistical physics methods such as Population Annealing (PA) \cite{Wang_2015}, which also employs a Gibbs-weighted ensemble to locate global ground states. However, while PA relies on discrete demographic selection, duplicating low-energy particles and eliminating high-energy ones prior to local Markov Chain Monte Carlo steps, our method maintains a fixed population size and drives exploration through continuous dynamical repulsion. This active exploration mechanism differentiates our Softmin Energy functional from the standard log-sum-exp (or free-energy) functional, $F_\beta(\mathbf{x}) = -{\beta}^{-1} \log \sum e^{-\beta f(x_i)}$. The gradient of the log-sum-exp formulation, $\nabla_{x_k} F_\beta(\mathbf{x}) = w_k \nabla f(x_k)$, strictly preserves gradient descent, causing sub-optimal particles with small weights to simply stagnate. In contrast, the gradient of our proposed Softmin functional, $J_\beta(\mathbf{x}) = \sum w_i f(x_i)$, natively incorporates the sign-changing effective learning rate, $\prefactor{t}{k}$. Consequently, rather than discarding trapped particles, the geometry of the Softmin functional actively destabilizes their local minima, forcing high-energy particles to hill-climb and escape via the topological barrier-flattening mechanism formalized in our theoretical analysis.

\paragraph{Softmin Energy} Theoretical work on the Gibbs Softmin (or Boltzmann Softmin) has primarily focused on its properties as a smooth approximation to the minimum operator \cite{bertsekas1997nonlinear}, convergence of softmax policy gradients in reinforcement learning \cite{mei2020global}, and its role in exploration–exploitation trade-offs via temperature scheduling \cite{song2019revisiting}. To our knowledge, this is the first work employing $J_\beta$ as an optimization objective for non-convex global search. The effective barrier reduction via particle cooperation, analyzed in Theorem \ref{thm:transition-time}, is previously unstudied.

\subsection{Structure of the paper}
In Section \ref{sec:gradient-flow-dynamics}, we analyze the deterministic gradient flow dynamics obtained by setting $\sigma =0$ in \eqref{e:sde} in detail, including the stability of stationary points. For a $\lambda$-strongly convex objective $f$ with $\lambda>0$, we show convergence to a stationary point where at least one particle reaches the global minimum of $f$. The stability analysis further reveals that particles outside this minimum exhibit exploratory behavior. 

In Section \ref{sec:hitting-times}, we examine the noisy dynamics \eqref{e:sde}, demonstrating that exploratory particles transition between local minima faster than an overdamped Brownian particle. This acceleration results from the interactions among particles, which reduce effective potential barriers. Moreover, particles within a local minimum's basin of attraction converge to it more rapidly than their overdamped counterparts.

In Section \ref{sec:numerical-experiments}, we conduct numerical experiments to validate the method's effectiveness in escaping local minima and identifying global optima. Comparisons with Simulated Annealing, Gradient-Aware CBO \cite{carrillo2021consensus}, and Gradient-Enhanced PSO \cite{noel2012new} show better performance, particularly when potential barriers exceed a critical threshold.

The proofs of lemmas and theorems throughout the paper are deferred to Appendix \ref{apx:technical-proofs}. Details on experimental settings can be found in Appendix \ref{app:hyperparameters}.

\section{Gradient Flow dynamics for Softmin Energy}
\label{sec:gradient-flow-dynamics}

This section focuses on the deterministic gradient flow dynamics
\begin{equation}
    d\mathbf{x}_t = -n\nabla J_\beta(\mathbf{x}_t)dt,
    \label{eq:gradient-flow}
\end{equation}
where $\mathbf{x}_t = (x_1(t), \dots, x_n(t))$ denotes the state of the particle system and the Softmin energy $J_\beta(\mathbf{x})$ is defined in \eqref{eq:Softmin-energy}. The scaling factor $n$, which can be absorbed via a time change, ensures that the subsequent derivatives and governing equations remain well-posed in the mean-field limit $n \to \infty$. This continuous-time system performs gradient descent on the Softmin landscape, which is formulated such that its minimization inherently drives the ensemble toward the minima of the objective function $f$.

Subsequent analysis establishes convergence guarantees under the assumption that $f$ is locally $\lambda$-strongly convex. Under this condition, the dynamics converge to a stationary configuration where at least one particle resides at a minimum of $f$. Furthermore, a rigorous stability analysis of these equilibria demonstrates that sub-optimal configurations, where specific particles remain at elevated function values, exhibit strict mathematical instability. This instability is driven by negative eigenvalues in the exact Hessian block, confirming that the Softmin formulation actively repels particles from sub-optimal critical points to promote landscape exploration.

To formally analyze these dynamics, it is necessary to first evaluate the gradient of the Softmin energy with respect to the individual particle coordinates. Let $w_k = e^{-\beta f(x_k)}(\sum_j e^{-\beta f(x_j)})^{-1}$ denote the Gibbs weight of particle $x_k$, and define the dynamic pre-factor $\prefactor{t}{k} = w_k\left[1 - \beta(f(x_k) - J_\beta(\mathbf{x}))\right]$. The gradient of $J_\beta(\mathbf{x})$ with respect to particle $x_k$ is computed in \ref{lem:derivative-of-J-proof} and is given by:
\begin{equation}
    \nabla_{x_k} J_\beta(\mathbf{x}) =\prefactor{t}{k} \nabla f(x_k).
    \label{eq:softmin-gradient}
\end{equation}
The exact second-order partial derivatives and corresponding Hessian blocks, which are required for the subsequent local stability analysis, are deferred to Appendix \ref{lem:derivative-of-J-proof}.

This formulation reveals a critical property of the Softmin gradient. The dynamic pre-factor $\prefactor{t}{k} = w_k\beta\left[1/\beta - f(x_k) + J_\beta(\mathbf{x})\right]$ dictates the effective direction of the gradient flow for each particle $x_k$, acting as a state-dependent learning rate. According to the dynamics in \eqref{eq:gradient-flow}, if a particle evaluates to a landscape value $f(x_k) < J_\beta(\mathbf{x}) + 1/\beta$, the pre-factor $\prefactor{t}{k}$ is strictly positive, and the particle performs standard gradient descent on $f$. Conversely, if $f(x_k) > J_\beta(\mathbf{x}) + 1/\beta$, the pre-factor becomes negative, effectively reversing the flow. In this regime, the particle executes gradient ascent on $f$. This mechanism theoretically guarantees that particles trapped in sub-optimal local minima are actively repelled and forced to explore the landscape, whereas particles at comparatively lower energy states continue to minimize $f$.

\subsection{Stationary points of the dynamics}
\label{sec:stationary-points}

This subsection characterizes the equilibria of the gradient flow \eqref{eq:gradient-flow} under the assumption that the objective function $f$ is locally $\lambda$-strongly convex \cite{bach2014adaptivity, fehrman2020convergence}:
\begin{definition}[Local Strong Convexity]
    \label{def:local-strong-convexity}
    A differentiable function $f:\mathbb{R}^d\to\mathbb{R}$ is said to be \textit{locally $\lambda$-strongly convex} if for every compact set $K\subset \text{dom}\nabla f$, there exists a constant $\lambda>0$ such that for all $x_1, x_2\in K$:
    \[
        f(x_1)\geq f(x_2)+\langle \nabla f(x_2), x_1-x_2\rangle + \frac{\lambda}{2}\|x_1-x_2\|^2.
    \]
\end{definition}
Theorem \ref{thm:stationary-points} establishes that the multi-particle system converges to a stationary configuration in which at least one particle resides at a minimum of $f$. Furthermore, the stability analysis shows that the pre-factor $\prefactor{t}{k}$ dictates two distinct dynamical regimes. While the equilibrium is asymptotically stable for particles located at the global or local minimum, sub-optimal particles experience a reversal of the local gradient field. This reversal induces a rigorous gradient ascent, actively repelling these particles from sub-optimal basins of attraction and enforcing landscape exploration. Consequently, the Softmin formulation inherently guarantees both the exploitation of deep minima and the systematic escape from higher-energy traps. Explicit convergence rates to these stationary configurations are derived subsequently in Section \ref{sec:hitting-times}.

\begin{theorem}
    \label{thm:stationary-points}
    Let $\mathbf{x}(t)$ denote the state of $n$ particles governed by the gradient flow \eqref{eq:gradient-flow}. Assume the objective function $f$ is locally $\lambda$-strongly convex and, without loss of generality, $\min f(x) = 0$. Then, the following properties hold:
    \begin{itemize}
        \item At any equilibrium configuration $\mathbf{\bar{x}} = (\bar{x}_1, \dots, \bar{x}_n)$, there exists an integer $1 \leq m \leq n$ such that $m$ particles reside at critical points of $f$ (i.e., $\nabla f(\bar{x}_i) = 0$), while the remaining $n-m$ particles satisfy the zero-pre-factor condition $f(\bar{x}_j) = J_\beta(\mathbf{\bar{x}}) + 1/\beta$ (equivalently, $\prefactor{t}{j} = 0$).
        \item The completely collapsed equilibrium ($m=n$), where all particles are located at the minimum of $f$, is locally asymptotically stable.
        \item For configurations where $m=n-1$, let $x^*$ denote the single exploratory particle satisfying $\nabla f(x^*) \neq 0$ and $\prefactor{t}{*} = 0$. This equilibrium is strictly unstable. Specifically, the diagonal block of the Hessian corresponding to $x^*$ possesses a strictly negative eigenvalue along the direction of $\nabla f(x^*)$.
    \end{itemize}
\end{theorem}

\begin{proof}
    The rigorous derivation of these stability conditions is deferred to Theorem \ref{thm:stationary-points-proof}.
\end{proof}

\begin{remark}
    \label{rem:local-convexity-topology}
    The assumption of local $\lambda$-strong convexity in Theorem \ref{thm:stationary-points} is introduced to model the topology of isolated basins of attraction within a generally non-convex landscape. Rather than restricting the global properties of the objective function $f$, this local geometric condition allows us to formalize the behavior of the system within specific traps. In particular, the theorem ensures that if the ensemble reaches the basin of the global optimum, the configuration acts as a stable attractor, guaranteeing convergence. Conversely, if the system is located in a suboptimal local minimum and at least one particle enters the maximizing regime (i.e., $\prefactor{t}{k} \leq 0$), the presence of a strictly negative eigenvalue guarantees that this particle will escape the basin. It is important to note, however, that if all particles were to simultaneously collapse into a suboptimal minimum, the deterministic gradient flow would remain trapped. This limitation naturally motivates the introduction of stochastic noise in the subsequent sections, which serves to break such complete suboptimal collapses and facilitate global exploration.
\end{remark}

\section{Hitting Times}
\label{sec:hitting-times}

This section analyzes the stochastic gradient flow dynamics, given by:
\begin{equation}
    d\mathbf{x}_t = -n\nabla J_\beta(\mathbf{x}_t)dt + \sqrt{2\sigma}dB_t,
    \label{eq:gradient-flow-noisy}
\end{equation}
where the term $\sqrt{2\sigma}dB_t$ introduces isotropic $nd$-dimensional Brownian noise. The subsequent analysis quantifies the theoretical advantages of the proposed Softmin formulation over standard Stochastic Gradient Descent (SGD), specifically evaluating both the expected transition times between isolated local minima and the convergence rates to these equilibria.

\subsection{Exit times from local minima}

This subsection establishes theoretical bounds on transition times between local minima in the small-noise regime. For an individual particle $x_k$, the stochastic dynamics in \eqref{eq:gradient-flow-noisy} can be expressed component-wise as:
\begin{equation}
    dx_k(t) = -\prefactor{t}{k}\nabla f(x_k(t))dt + \sqrt{2\sigma}dB_t.
    \label{eq:softmin-langevin}
\end{equation}
Here, $B_t$ is a $d$-dimensional Brownian noise, and $\prefactor{t}{k}$ acts as a state-dependent effective learning rate for particle $x_k$, explicitly given by:
\begin{small}
\begin{equation}
    \prefactor{t}{k} = \beta\left[\frac{1}{\beta} - f(x_k(t)) + J_\beta(\mathbf{x}_t)\right]\frac{ne^{-\beta f(x_k(t))}}{\sum_{j=1}^n e^{-\beta f(x_j(t))}}.
    \label{eq:pre-factor}
\end{equation}
\end{small}
Assuming without loss of generality that $\min f(x)=0$, we classify the behavioral regime of particle $x_k$ over a given time interval $I$ based on the sign of this effective rate. The particle is \textit{minimizing} when it possesses a positive effective learning rate ($\prefactor{t}{k} > 0$ for all $t$ on some time interval $I$, thus performing standard gradient descent). Conversely, the particle is \textit{maximizing} when this rate becomes negative ($\prefactor{t}{k} < 0$, inherently reversing the flow to perform gradient ascent). 

To rigorously estimate these transition times, the stochastic gradient flow of the entire ensemble \eqref{eq:gradient-flow-noisy} must be treated as a single Langevin diffusion in $\mathbb{R}^{nd}$, driven by the joint potential $U(\mathbf{x}) = n J_\beta(\mathbf{x})$. Let $D$ be the basin of attraction of $x_{\min}$, and assume the system is initialized at the joint equilibrium $\bar x := (x_{\min},\ldots,x_{\min})\in \mathbb{R}^{nd}$. Let $x^\ast\in\partial D$ be the relevant index-one saddle separating $D$ from the adjacent basin, with physical barrier $f^\ast := f(x^\ast)$. 

By the Freidlin-Wentzell Large Deviation Principle (LDP), the exponential scale of the exit time is governed by the minimum action path required to reach the boundary $\partial D$. For $q \in \{1, \dots, n\}$, let $V_q$ denote the minimum Freidlin-Wentzell action required for exactly $q$ particles to reach $\partial D$ while the remaining components stay within $D$. In particular, $V_1$ is the exact one-particle Freidlin-Wentzell quasipotential, corresponding to the configuration where a single particle reaches $\partial D$, while the other $n-1$ particles stay fixed at $x_{\min}$. We define the explicit Softmin barrier bound $\overline{V}_\beta$ as:
\begin{equation}
    \overline{V}_\beta := \max\left\{ \frac{n}{2(n-1)\beta}, \frac{\beta}{2}\left(f^\ast-\frac1\beta\right)_+^2 \right\}, \qquad (a)_+ := \max\{a,0\}.
\end{equation}

The following theorem demonstrates that the Softmin formulation penalizes multi-particle transitions, proving that the exact transition time scales with the single-particle action $V_1$, which is analytically bounded by $\overline{V}_\beta$.

\begin{theorem}
\label{thm:transition-time}
    Assume that $f\in C^3$ is Morse and coercive, $x_{\min}$ is a non-degenerate local minimum, and $x^\ast$ is the minimal saddle point on the boundary $\partial D$. Furthermore, assume that the relevant exit points of $nJ_\beta$ are isolated, non-degenerate saddle points. Then there exists $n_0\in\mathbb{N}$ such that, for all $n\ge n_0$, the expected transition time for any particle $k$ scales logarithmically as:
    \[
        \lim_{\sigma\downarrow0} \sigma\log \mathbb{E}_{\bar x}\tau_k^\sigma = V_1.
    \]
    Moreover, the probability that exactly one particle exits $D$ during the transition tends to $1$ as $\sigma\downarrow0$. Furthermore, the exact exponential scale is bounded by the computable Softmin barrier:
    \[
        V_1\le \overline{V}_\beta.
    \]
    Consequently, for every $\eta>0$, there exists $\sigma_0>0$ such that, for all $0<\sigma<\sigma_0$, the expected transition time satisfies 
    \[
        \mathbb{E}_{\bar x}\tau_k^\sigma \le \exp\left(\frac{\overline{V}_\beta+\eta}{\sigma}\right).
    \]
\end{theorem}

\begin{proof}
The full mathematical proof is deferred to Appendix \ref{thm:transition-time-proof}. 

The stochastic transition time is exponentially governed by the Freidlin-Wentzell rate functional. We first establish channel separation (Lemma \ref{lem:single-particle-channel}): by evaluating the zero-prefactor condition at the relevant saddle points, we prove that multi-particle saddles ($V_q$ for $q \ge 2$) have strictly larger action than the one-particle saddle $V_1$. Thus, the probability of multi-particle transitions is exponentially suppressed, and the exact escape scale is $V_1$. 

To compute the upper bound $\overline{V}_\beta$, we evaluate a specific one-dimensional test channel where $n-1$ particles are frozen at $x_{\min}$ while a single active particle escapes. Since $V_1$ is an infimum over all valid paths, the action of this specific frozen-background channel provides a strict upper bound. This 1D channel structurally decomposes into two bottlenecks:
\begin{itemize}
    \item \textbf{Ejection:} Exiting the local minimum is aided by Softmin repulsion. Bounding the active particle's Gibbs weight yields a linear majorant for the effective learning rate. Integrating this majorant caps the ejection action at $n/(2(n-1)\beta)$.
    \item \textbf{Descent:} Approaching the physical peak $f^*$, the repulsive dynamics invert, creating an opposing trap. Integrating the maximal opposing gradient provides a strict bound on the descent action of $(\beta/2)(f^* - 1/\beta)_+^2$.
\end{itemize}
Taking the maximum of these two sequential bottlenecks yields $V_1 \le \overline{V}_\beta$. Finally, standard Freidlin-Wentzell confinement estimates guarantee that the non-escaping particles remain localized in an arbitrarily small neighborhood of $x_{\min}$ during the transition window with probability tending to one.
\end{proof}

\begin{remark}[Acceleration over Naive Langevin Dynamics] 
    Compared to a naive Langevin process (standard noisy gradient descent) where the exact barrier is the physical peak $\Delta V_{\text{naive}} = f^*$, the Softmin dynamics strictly accelerate the expected transition time if the inverse temperature $\beta$ is tuned such that the computable upper bound is smaller than $f^*$. For a sufficiently high barrier where the descent phase dominates ($f^* > 1/\beta$), enforcing $\overline{V}_\beta < f^*$ yields the strict condition $\beta < (2 + \sqrt{3})/f^*$. Under this condition, the expected stochastic escape is guaranteed to be exponentially faster.
\end{remark}


\begin{remark}[Dynamic Annealing of the Inverse Temperature $\beta$]
    The bounding condition $\beta < (2+\sqrt{3})/f^*$ exposes a fundamental trade-off. A large $\beta$ is advantageous for ejection, minimizing the first term of $\overline{V}_\beta \propto 1/\beta$. Conversely, a large $\beta$ deepens the dynamically induced trap at the peak, maximizing the second term $\propto (\beta/2)(f^* - 1/\beta)^2$. This tension rigorously justifies a decaying annealing schedule for $\beta$. Initializing with a high $\beta$ aggressively flattens the initial basin. Decaying $\beta \to 0$ systematically pushes the transition point $1/\beta \to \infty$. Once $1/\beta \ge f^*$, the trapped maximizing particles revert to a minimizing regime, escaping the local maximum to descend into the basins of attraction of adjacent minima.
\end{remark}

\subsection{Convergence rate to local minima}

For a particle operating within the minimizing regime, we seek to compare its convergence rate to a local minimum against that of standard gradient descent (an overdamped Brownian particle). A priori, it is not guaranteed even in the deterministic setting (Eq. \ref{eq:gradient-flow}) that a minimizing particle remains strictly minimizing throughout its trajectory. 

Recall that a particle $k$ is defined as \textit{minimizing} over a time interval $I$ if its effective learning rate is strictly positive, $\prefactor{t}{k}>0$, for all $t\in I$. We introduce a stricter condition: a particle is \textit{strongly minimizing} over $I$ if its objective value satisfies $f(x_k(t))\leq J_\beta(\mathbf{x}_t)$ for all $t\in I$. By the definition of the Softmin dynamics, this strongly minimizing condition strictly implies an accelerated learning rate, $\prefactor{t}{k} \geq 1$. 

We first establish that these properties are invariant under the deterministic flow.

\begin{lemma}
    \label{lem:minimizing-particle}
    If a particle $x_k$ is minimizing (resp., strongly minimizing) at time $t=0$, then under the deterministic gradient flow dynamics (Eq. \ref{eq:gradient-flow}), it remains minimizing (resp., strongly minimizing) for all $t>0$.
\end{lemma}
\begin{proof}
    The proof is deferred to Lemma \ref{lem:minimizing-particle-proof}.
\end{proof}

Now we are ready to compute convergence rates for the minimizing particles in the deterministic dynamics in Equation \eqref{eq:gradient-flow}. 

\begin{theorem}
    \label{thm:rate-of-convergence-minimizing-particle}
    Let $f$ be locally strongly convex and $x_k(0)$ be a strongly minimizing particle, so that $\prefactor{0}{k}>1$. Assume that $x_k(0),x^*\in K$ where $x^*\in \argmin f$ and $K$ is compact. Then, the deterministic dynamics in \eqref{eq:gradient-flow} converges to $x^*$ at a rate $\mathcal{O}(e^{-\lambda t})$, where $\lambda$ is the strong convexity constant of the function $f$ in the compact set $K$.
\end{theorem}
\begin{proof}
    The proof is deferred to Theorem \ref{thm:rate-of-convergence-minimizing-particle-proof}.
\end{proof}

\begin{remark}
    If we consider the noisy dynamics in Equation \eqref{eq:gradient-flow-noisy}, and the particle is minimizing but not strongly minimizing, then the noise term will dominate over the gradient term in the dynamics. In this case, we can consider the dynamics as driven by noise and the particle can either transition to the maximizing regime or stay in the minimizing regime and become strongly minimizing. The probabilities for these two cases can be estimated by considering the particle to be only subject to noise, and will depend on the steepness of the function $f$ at the point $x_k$. This technical computation  is left for future work.
\end{remark}

In summary, we have established that the Softmin dynamics systematically modulate the landscape of the objective function $f$. Specifically, we demonstrated that by selecting an inverse temperature $\beta < (2+\sqrt{3})/f^*$, the dynamics effectively lower the potential barriers, thereby strictly reducing the expected transition times between local minima in the small-noise regime ($\sigma \to 0$). Furthermore, for particles in the strongly minimizing regime ($\prefactor{t}{k} \geq 1$), we proved that the deterministic convergence rate is at least as fast as that of standard gradient descent, and strictly faster when $\prefactor{t}{k} > 1$. These theoretical results indicate that Softmin Langevin dynamics simultaneously accelerate both global exploration and local refinement. In the following section, we provide empirical validation of these results through a series of numerical experiments on non-convex landscapes.

\section{Numerical Experiments}
\label{sec:numerical-experiments}

We evaluate the capacity of Softmin dynamics to escape local minima by comparing fixed ($\beta=10$) and decaying schedules against Simulated Annealing (SA), Gradient-Aware Consensus-Based Optimization (CBO) \cite{carrillo2021consensus}, and Gradient-Enhanced Particle Swarm Optimization (PSO) \cite{noel2012new}. The continuous dynamics are discretized using the Euler-Maruyama scheme. Landscape formulations, hyperparameters, initialization bounds, and hardware specifications are detailed in Appendix \ref{app:hyperparameters}; the source code is available at \cite{softmin_energy_2026}. Because all evaluated methods require $O(n)$ gradient computations per step, they share an identical per-iteration computational cost. Consequently, the advantage of Softmin lies not in cheaper local updates, but in exponentially shorter transition times across topological barriers.

We first verify the large deviation scaling of Theorem \ref{thm:transition-time} on a 1D double well ($\sigma = 0.05$). To isolate the exit phase, all particles are initialized in the left-most minimum. Results are shown in Appendix \ref{apx:other-experiments}. 

\begin{figure*}[!htbp]
  \centering
  \begin{minipage}{0.32\textwidth}
    \centering
    \includegraphics[width=\linewidth]{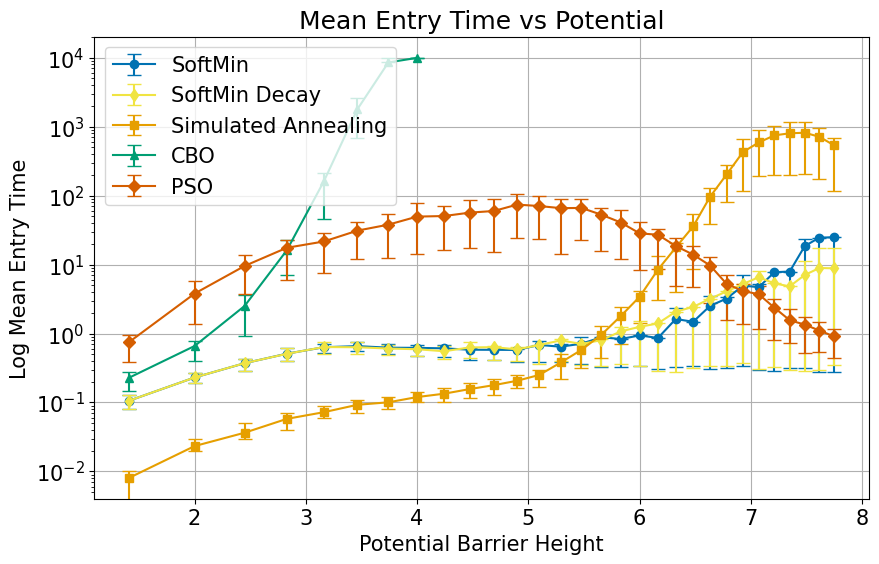}
  \end{minipage}\hfill
  \begin{minipage}{0.32\textwidth}
    \centering
    \includegraphics[width=\linewidth]{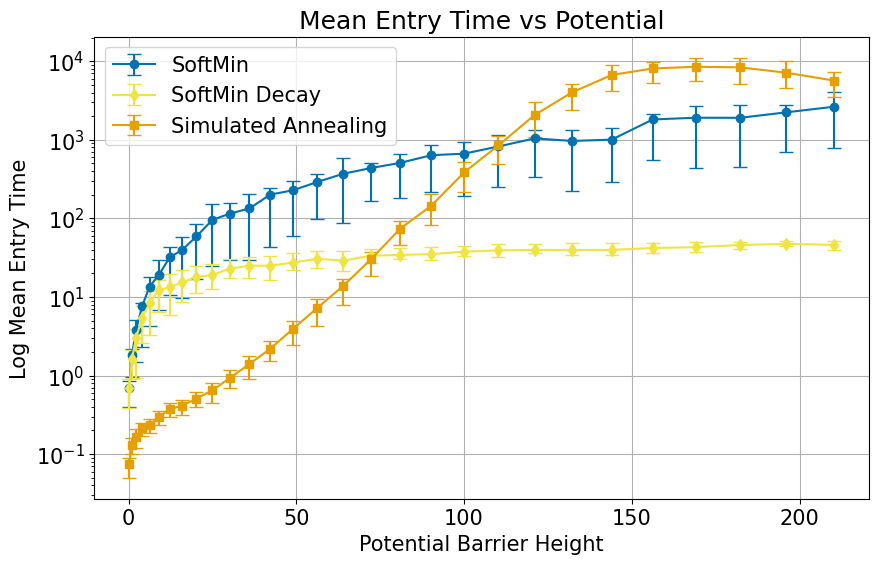}
  \end{minipage}\hfill
  \begin{minipage}{0.32\textwidth}
    \centering
    \includegraphics[width=\linewidth]{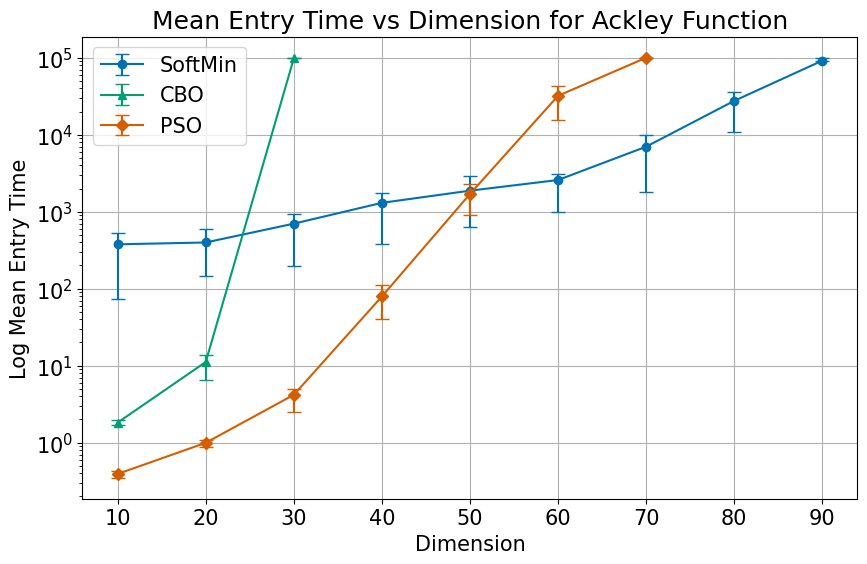}
  \end{minipage}

  \vspace{\baselineskip}
  \caption{Numerical validation of the Softmin dynamics. Left: 1d double well benchmark. Center: 2d quadruple well benchmark (CBO and PSO omitted due to complete variance collapse). Right: Dimensional scalability on the generalized Ackley function, max steps allowed $10^5$ (SA omitted due to maximum steps reached at $d=10$). The error bars in the plots are drawn in correspondence to the quartiles of the distribution of the computed exit times.}
  \label{fig:experiments}
  \vspace{\baselineskip}
\end{figure*}

To evaluate topological escape ($\sigma=1$), subsequent benchmarks initialize the swarm entirely within a suboptimal basin to test each method's capacity to generate exploratory variance before reaching the maximum step limit. Figure \ref{fig:experiments} (Left) reports the 1D performance across varying barrier heights $h$. While Softmin variants demonstrate a stable, flat trajectory, CBO fails due to premature consensus, and SA exhibits severe exponential scaling. At extreme barriers ($h > 6.5$), SA and PSO display anomalous non-monotonicity due to a numerical discretization artifact: steep outer walls induce large discrete gradient steps, causing SA to artificially overshoot the barrier and PSO to escape via inertia. Softmin prevents this numerical instability through its intrinsically bounded pre-factor $\prefactor{t}{k}$.

Figure \ref{fig:experiments} (Center) extends this analysis to an unbalanced 2D quadruple well. In this higher-dimensional setting, the numerical artifacts observed in 1D vanish; particles subjected to large discrete gradient updates typically scatter off steep walls rather than traversing the narrow topological saddle. Consequently, both PSO and CBO fail to escape the suboptimal well due to variance collapse and are omitted from the plot. Simulated Annealing again exhibits exponential scaling with barrier height, eventually plateauing when the required transition time exceeds the maximum step limit. In contrast, the Softmin methods outperform baseline techniques by actively repelling particles to generate the spatial variance necessary for escape. Furthermore, the advantage of the geometrically decaying schedule is evident: while fixed Softmin ($\beta=10$) exhibits a slight upward trend at extreme barriers, the decaying variant maintains consistently lower transition times.

Finally, Figure \ref{fig:experiments} (Right) evaluates scalability on the generalized Ackley function across 10 to 90 dimensions. Particles are initialized uniformly over a compact domain containing an exponential number of local minima. SA fails to converge across all tested dimensions. As the search space volume increases, CBO and PSO reach the computational step limit without escaping by dimensions 30 and 70, respectively. In contrast, Softmin scales robustly. By actively repelling particles from sub-optimal basins, the dynamics effectively drive the swarm toward the global optimum, demonstrating its efficacy in complex, high-dimensional landscapes.

\section{Conclusion}

In this work, we introduced a novel method for escaping local minima in optimization problems, based on the Softmin Energy interaction dynamics. We showed that the dynamics is able to escape local minima and find global optima by lowering effective potential barriers, leading to faster transition times between local minima in the small noise regime. We also proved that the convergence rate of the dynamics is faster than the one of a particle following the Langevin dynamics. Through numerical experiments, we validated our theoretical results and demonstrated the effectiveness of our method on benchmark functions such as double wells and the Ackley function. In the analyzed benchmarks, our method outperforms Simulated Annealing, Gradient-Enhanced Particle Swarm Optimization, and Gradient-Aware Consensus-Based Optimization in terms of transition times and convergence rates.

\subsection{Limitations and Future Works} 
Our theoretical analysis provides rigorous guarantees in the practically relevant small-noise regime where the Friedlin-Wentzell Large Deviation Principles apply. While the exponential bound in Theorem \ref{thm:transition-time} is not sharp, leaving room for a tighter characterization of the exponential rate in the regime where $\overline{V}_\beta$ scales quadratically in $f^*$, together with intermediate-noise asymptotics, the framework establishes fundamental advantages of particle cooperation for barrier reduction. Our convergence analysis (Theorems \ref{thm:stationary-points} and \ref{thm:rate-of-convergence-minimizing-particle}) leverages strong convexity to yield explicit rates; extending these guarantees to non-convex objectives is technically challenging because the minimizer/maximizer partition (Theorem \ref{thm:stationary-points}) relies critically on the gradient being zero only at isolated minima, a property that breaks down when multiple critical points compete.

Future work should address several important directions. First, extending convergence proofs and explicit convergence rates to the non-convex regime, particularly for objectives with multiple disconnected basins, would significantly broaden applicability. Second, analyzing the mean-field limit as $n \to \infty$ could provide insights into the scaling of the method with ensemble size and potentially yield dimension-free convergence rates. Finally, tighter estimates for the exponential rate in the small noise regime can sharpen the theoretical bound to match experimental evidence more closely.

\bibliographystyle{plain}  
\bibliography{references}  

\newpage
\appendix
\onecolumn
\section{Technical Proofs}
\label{apx:technical-proofs}

Here we will collect some of the more technical proofs of the lemmas stated in the main body. 

\subsection{Proofs of Section \ref{sec:gradient-flow-dynamics}}

\begin{lemma}
    \label{lem:derivative-of-J-proof}
    Let $w_k = e^{-\beta f(x_k)}(\sum_j e^{-\beta f(x_j)})^{-1}$ denote the Gibbs weight of particle $x_k$, and define the dynamic pre-factor $\prefactor{t}{k} = w_k\left[1 - \beta(f(x_k) - J_\beta(\mathbf{x}))\right]$. The gradient of $J_\beta(\mathbf{x})$ with respect to particle $x_k$ is given by:
    \begin{equation*}
        \nabla_{x_k} J_\beta(\mathbf{x}) = \prefactor{t}{k} \nabla f(x_k).
    \end{equation*}
    The mixed second partial derivative of $J_\beta(\mathbf{x})$ with respect to particles $x_k$ and $x_\ell$ (for $k \neq l$) is:
    \begin{equation*}
        \frac{\partial^2 J_\beta(\mathbf{x})}{\partial x_k \partial x_\ell} = \beta^2 w_k w_\ell \left[ \frac{2}{\beta} - f(x_k) - f(x_\ell) + 2J_\beta(\mathbf{x}) \right] \nabla f(x_k) \nabla f(x_\ell)^\top.
    \end{equation*}
    Furthermore, the diagonal block of the Hessian with respect to particle $x_k$ is given by:
    \begin{equation*}
        \frac{\partial^2 J_\beta(\mathbf{x})}{\partial x_k^2} = \prefactor{t}{k} \nabla^2 f(x_k) - \beta \left[ w_k + \prefactor{t}{k} - 2 w_k \prefactor{t}{k} \right] \nabla f(x_k) \nabla f(x_k)^\top.
    \end{equation*}
\end{lemma}

\begin{proof}
    We compute the derivatives using the chain rule and the properties of the Gibbs weights. First, we establish the gradient of the weight $w_i$ with respect to $x_k$:
    \begin{equation*}
        \nabla_{x_k} w_i = \begin{cases}
            \beta w_i w_k \nabla f(x_k) & \text{if } i \neq k, \\
            -\beta w_k (1 - w_k) \nabla f(x_k) & \text{if } i = k.
        \end{cases}
    \end{equation*}
    For the first derivative of the Softmin energy $J_\beta(\mathbf{x}) = \sum_i w_i f(x_i)$, we obtain:
    \begin{align*}
        \nabla_{x_k} J_\beta(\mathbf{x}) &= \sum_{i \neq k} (\nabla_{x_k} w_i) f(x_i) + (\nabla_{x_k} w_k) f(x_k) + w_k \nabla f(x_k) \\
        &= \sum_{i \neq k} \beta w_i w_k f(x_i) \nabla f(x_k) - \beta w_k (1 - w_k) f(x_k) \nabla f(x_k) + w_k \nabla f(x_k) \\
        &= \beta w_k \left[ \sum_i w_i f(x_i) - w_k f(x_k) - (1 - w_k) f(x_k) \right] \nabla f(x_k) + w_k \nabla f(x_k) \\
        &= \beta w_k \left[ J_\beta(\mathbf{x}) - f(x_k) \right] \nabla f(x_k) + w_k \nabla f(x_k) \\
        &= w_k \left[ 1 - \beta (f(x_k) - J_\beta(\mathbf{x})) \right] \nabla f(x_k) = \prefactor{t}{k} \nabla f(x_k).
    \end{align*}
    
    For the mixed second derivative ($k \neq l$), we differentiate $\nabla_{x_k} J_\beta(\mathbf{x})$ with respect to $x_\ell$. Because $\nabla f(x_k)$ is independent of $x_\ell$, the Jacobian matrix is derived via the outer product:
    \begin{align*}
        \frac{\partial^2 J_\beta(\mathbf{x})}{\partial x_k \partial x_\ell} &= \nabla f(x_k) \left( \nabla_{x_\ell} \left[ \prefactor{t}{k} \right] \right)^\top \\
        &= \nabla f(x_k) \left[ (\nabla_{x_\ell} w_k) \left[1 - \beta(f(x_k) - J_\beta(\mathbf{x}))\right] + w_k (\nabla_{x_\ell} \left[1 - \beta(f(x_k) - J_\beta(\mathbf{x}))\right]) \right]^\top.
    \end{align*}
    Substituting $\nabla_{x_\ell} w_k = \beta w_k w_\ell \nabla f(x_\ell)$ and $\nabla_{x_\ell} \left[1 - \beta(f(x_k) - J_\beta(\mathbf{x}))\right] = \beta \nabla_{x_\ell} J_\beta(\mathbf{x}) = \beta \prefactor{t}{\ell} \nabla f(x_\ell)$ into the bracket yields:
    \begin{align*}
        \frac{\partial^2 J_\beta(\mathbf{x})}{\partial x_k \partial x_\ell} &= \nabla f(x_k) \left[ \beta w_\ell \prefactor{t}{k} \nabla f(x_\ell) + \beta w_k \prefactor{t}{\ell} \nabla f(x_\ell) \right]^\top \\
        &= \beta w_k w_\ell \left[ 2 - \beta(f(x_k) + f(x_\ell) - 2J_\beta(\mathbf{x})) \right] \nabla f(x_k) \nabla f(x_\ell)^\top \\
        &= \beta^2 w_k w_\ell \left[ \frac{2}{\beta} - f(x_k) - f(x_\ell) + 2J_\beta(\mathbf{x}) \right] \nabla f(x_k) \nabla f(x_\ell)^\top.
    \end{align*}

    Finally, for the diagonal block ($k = l$), we differentiate $\nabla_{x_k} J_\beta(\mathbf{x})$ with respect to $x_k$. Applying the product rule includes the local Hessian $\nabla^2 f(x_k)$:
    \begin{align*}
        \frac{\partial^2 J_\beta(\mathbf{x})}{\partial x_k^2} &=  \prefactor{t}{k} \nabla^2 f(x_k) + \nabla f(x_k) \left( \nabla_{x_k} \left[  \prefactor{t}{k} \right] \right)^\top \\
        &= \prefactor{t}{k} \nabla^2 f(x_k) + \nabla f(x_k) \big[ (\nabla_{x_k} w_k) \left[1 - \beta(f(x_k) - J_\beta(\mathbf{x}))\right]\\
        &\qquad + w_k (\nabla_{x_k} \left[1 - \beta(f(x_k) - J_\beta(\mathbf{x}))\right]) \big]^\top.
    \end{align*}
    Using $\nabla_{x_k} w_k = -\beta w_k (1 - w_k) \nabla f(x_k)$ and $\nabla_{x_k} \left[1 - \beta(f(x_k) - J_\beta(\mathbf{x}))\right] = -\beta \nabla f(x_k) + \beta \prefactor{t}{k} \nabla f(x_k) = -\beta (1 - \prefactor{t}{k}) \nabla f(x_k)$, the gradient of the pre-factor $\prefactor{t}{k}$ evaluates to:
    \begin{align*}
        \nabla_{x_k} \prefactor{t}{k} &= -\beta (1 - w_k) \prefactor{t}{k} \nabla f(x_k) - \beta w_k(1 - \prefactor{t}{k}) \nabla f(x_k) \\
        &= -\beta \left[ (1 - w_k) \prefactor{t}{k} + w_k(1 - \prefactor{t}{k}) \right] \nabla f(x_k) \\
        &= -\beta \left[ \prefactor{t}{k} - w_k \prefactor{t}{k} + w_k - w_k \prefactor{t}{k} \right] \nabla f(x_k) \\
        &= -\beta \left[ w_k + \prefactor{t}{k} - 2w_k \prefactor{t}{k} \right] \nabla f(x_k).
    \end{align*}
    Substituting this expression back into the product rule yields the final diagonal Hessian block:
    \begin{equation*}
        \frac{\partial^2 J_\beta(\mathbf{x})}{\partial x_k^2} = \prefactor{t}{k} \nabla^2 f(x_k) - \beta \left[ w_k + \prefactor{t}{k} - 2 w_k \prefactor{t}{k} \right] \nabla f(x_k) \nabla f(x_k)^\top. \qedhere
    \end{equation*}
\end{proof}

\begin{theorem}
    \label{thm:stationary-points-proof}
    Let $\mathbf{x}(t)$ denote the state of $n$ particles governed by the gradient flow \eqref{eq:gradient-flow}. Assume the objective function $f$ is locally $\lambda$-strongly convex and, without loss of generality, $\min f(x) = 0$. Then, the following properties hold:
    \begin{itemize}
        \item At any equilibrium configuration $\mathbf{\bar{x}} = (\bar{x}_1, \dots, \bar{x}_n)$, there exists an integer $1 \leq m \leq n$ such that $m$ particles reside at critical points of $f$ (i.e., $\nabla f(\bar{x}_i) = 0$), while the remaining $n-m$ particles satisfy the zero-pre-factor condition $\prefactor{t}{j} = 0$ (which, because $w_j > 0$, is equivalent to $f(\bar{x}_j) = J_\beta(\mathbf{\bar{x}}) + 1/\beta$).
        \item The completely collapsed equilibrium ($m=n$), where all particles are located at the minimum of $f$, is locally asymptotically stable.
        \item For configurations where $m=n-1$, let $x^*$ denote the single exploratory particle satisfying $\nabla f(x^*) \neq 0$ and $\prefactor{t}{*} = 0$. This equilibrium is strictly unstable. Specifically, the diagonal block of the Hessian corresponding to $x^*$ possesses a strictly negative eigenvalue along the direction of $\nabla f(x^*)$.
    \end{itemize}
\end{theorem}

\begin{proof}
    From Lemma \ref{lem:derivative-of-J-proof}, the gradient of the Softmin functional with respect to particle $x_k$ is given by $\nabla_{x_k} J_\beta(\mathbf{x}) = \prefactor{t}{k} \nabla f(x_k)$, where the dynamic pre-factor is defined as $\prefactor{t}{k} = w_k \left[1 - \beta(f(x_k) - J_\beta(\mathbf{x}))\right]$. A stationary point requires $\nabla_{x_k} J_\beta(\mathbf{x}) = 0$ for all $k \in \{1, \dots, n\}$. Because the Gibbs weights strictly satisfy $w_k > 0$, each particle must satisfy either $\nabla f(x_k) = 0$ or $\prefactor{t}{k} = 0$. The latter condition implies $1 - \beta(f(x_k) - J_\beta(\mathbf{x})) = 0$, which simplifies to $f(x_k) = J_\beta(\mathbf{x}) + 1/\beta$. 

    Assume by contradiction that all $n$ particles satisfy $\prefactor{t}{k} = 0$. Multiplying the equivalent condition $f(x_k) = J_\beta(\mathbf{x}) + 1/\beta$ by $w_k$ and summing over all $k$ yields $\sum_{k=1}^n w_k f(x_k) = \sum_{k=1}^n w_k \left(J_\beta(\mathbf{x}) + 1/\beta\right)$. By the definition of $J_\beta(\mathbf{x})$, this simplifies to $J_\beta(\mathbf{x}) = J_\beta(\mathbf{x}) + 1/\beta$, which implies $0 = 1/\beta$. Because $\beta < \infty$, this is a contradiction. Therefore, at least one particle must satisfy $\nabla f(x_k) = 0$, establishing $m \geq 1$.

    For the fully collapsed equilibrium ($m=n$), all particles satisfy $\nabla f(x_k) = 0$. Assume they reach a consensus at the same minimum such that $f(x_k) = \bar{f}$ for all $k$. Then, $J_\beta(\mathbf{x}) = \bar{f}$, the Gibbs weights are uniform $w_k = 1/n$, and consequently, the pre-factors evaluate to $\prefactor{t}{k} = \frac{1}{n}[1 - \beta(\bar{f} - \bar{f})] = 1/n$. Evaluating the Hessian derived in Lemma \ref{lem:derivative-of-J-proof}, the off-diagonal blocks vanish identically because $\nabla f(x_k) = 0$. The diagonal blocks simplify to:
    \[
        \frac{\partial^2 J_\beta(\mathbf{x})}{\partial x_k^2} = \prefactor{t}{k} \nabla^2 f(x_k) - 0 = \frac{1}{n} \nabla^2 f(x_k).
    \]
    Because $f$ is locally strongly convex at this minimum, $\nabla^2 f(x_k) \succ 0$. The full block-diagonal Hessian is therefore positive definite, confirming the completely collapsed equilibrium is locally asymptotically stable.

    For the configuration where $m=n-1$, let $x^*$ be the single exploring particle satisfying $\nabla f(x^*) \neq 0$. For this particle, the stationarity condition dictates $\prefactor{t}{*} = 0$. Evaluating the diagonal block of the Hessian with respect to $x^*$ using the formulation from Lemma \ref{lem:derivative-of-J-proof} yields:
    \[
        \frac{\partial^2 J_\beta(\mathbf{x})}{\partial (x^*)^2} = \prefactor{t}{*} \nabla^2 f(x^*) - \beta \left[ w^* + \prefactor{t}{*} - 2 w^* \prefactor{t}{*} \right] \nabla f(x^*) \nabla f(x^*)^\top
    \]
    Substituting $\prefactor{t}{*} = 0$, the term involving the local Hessian $\nabla^2 f(x^*)$ vanishes completely, resulting in:
    \[
        \frac{\partial^2 J_\beta(\mathbf{x})}{\partial (x^*)^2} = -\beta w^* \nabla f(x^*) \nabla f(x^*)^\top 
    \]
    This matrix is rank-one and negative semi-definite. Because $w^* > 0$, $\beta > 0$, and $\nabla f(x^*) \neq 0$, it possesses exactly one strictly negative eigenvalue, explicitly given by $-\beta w^* \|\nabla f(x^*)\|^2$, with the corresponding eigenvector aligned precisely with $\nabla f(x^*)$. The orthogonal subspace exhibits flat curvature (zero eigenvalues). The existence of this strictly negative eigenvalue mathematically guarantees that the equilibrium is unstable. The particle $x^*$ is actively repelled along the axis of the local gradient, destabilizing the sub-optimal configuration.
\end{proof}

\subsection{Proofs of Section \ref{sec:hitting-times}}
The following definitions and lemmas provide the rigorous Freidlin-Wentzell Large Deviation Principle (LDP) framework required to prove Theorem \ref{thm:transition-time}. 

For an absolutely continuous path $\phi:[0,T]\to\mathbb{R}^{nd}$, the Freidlin-Wentzell rate functional for the joint system is:
\[
    I_{0T}(\phi) := \frac14 \int_0^T \left\| \dot \phi(t)+\nabla(nJ_\beta)(\phi(t)) \right\|^2\dd t.
\]
For $q\in\{1,\ldots,n\}$, let $\mathcal{A}_q(T)$ be the class of absolutely continuous paths $\phi:[0,T]\to\mathbb{R}^{nd}$ such that $\phi(0)=\bar x$ and exactly $q$ particle components reach $\partial D$ by time $T$, while the other components remain in $D$. The minimum action is defined as $V_q := \inf_{T>0}\inf_{\phi\in\mathcal{A}_q(T)} I_{0T}(\phi)$.

\begin{lemma}
\label{lem:single-particle-channel}
    Under the assumptions of Theorem \ref{thm:transition-time}, the one-particle channel satisfies $V_1\le \overline{V}_\beta$. Moreover, there exists $n_0\in\mathbb{N}$ such that, for all $n\ge n_0$, the multi-particle channels are strictly separated:
    \[
        V_q > V_1, \qquad q=2,\ldots,n.
    \]
\end{lemma}

\begin{proof}
    We first bound the one-particle channel. Along the frozen-background path, keep $n-1$ particles at $x_{\min}$ and let the active particle have objective value $r=f(x_k)\ge 0$. Then $w(r)=e^{-\beta r}/(n-1+e^{-\beta r})$, and the effective coefficient along this channel is $A(r)=n w(r)\left[1-\beta(1-w(r))r\right]$. The associated one-dimensional effective potential satisfies $U_{\mathrm{eff}}'(r)=A(r)$.
    
    Since $V_1$ is an infimum over admissible one-particle paths, the cost of this specific channel gives an upper bound for $V_1$. For the ejection barrier, $w(r)\le 1/n$ and hence $1-w(r)\ge (n-1)/n$. On the interval where $A(r)\ge0$, we bound $A(r)\le 1-r\beta(n-1)/n$. The right-hand side vanishes at $r_{\max}=n/(\beta(n-1))$. Therefore, the ejection cost is bounded by:
    \[
        V_{\mathrm{exit}} \le \int_0^{r_{\max}} \left(1-\beta\frac{n-1}{n}s\right)\dd s = \frac{n}{2(n-1)\beta}.
    \]
    For the descent barrier, the opposing part of the drift satisfies $[-A(r)]_+ = n w(r)\left[\beta(1-w(r))r-1\right]_+ < (\beta r-1)_+$, because $nw(r)\le1$ and $1-w(r)<1$. Thus:
    \[
        V_{\mathrm{descent}} < \int_{1/\beta}^{f^\ast}(\beta s-1)\dd s = \frac{\beta}{2}\left(f^\ast-\frac1\beta\right)_+^2.
    \]
    Taking the maximum yields $V_1 \le \max\{V_{\mathrm{exit}},V_{\mathrm{descent}}\} \le \overline{V}_\beta$.

    We now prove separation of the multi-particle channels. Fix $1\le q\le n-1$. Consider a local $q$-particle Softmin saddle in which $q$ active particles have common objective value $r>0$, and the remaining $n-q$ particles remain at $x_{\min}$. Along this channel, $J_q(r) = (q r e^{-\beta r})/(n-q+q e^{-\beta r})$, with joint potential $U_q(r):=nJ_q(r)$.
    
    At a Softmin saddle for the active particles, the zero-prefactor condition gives $r-J_q(r)=\frac1\beta$. Substituting $y=\beta r$ and $\alpha_q=q/(n-q)$, this equation becomes $y=1+\alpha_q e^{-y}$. Let $y_q$ be its unique solution. Since $\alpha_q$ is strictly increasing in $q$ and $\frac{\dd y_q}{\dd \alpha_q} = \frac{e^{-y_q}}{1+\alpha_q e^{-y_q}}>0$, the map $q\mapsto y_q$ is strictly increasing.
    
    At the saddle, $U_q(r_q) = n\left(r_q-1/\beta\right) = (n/\beta)(y_q-1)$. Therefore, $U_1(r_1)<U_2(r_2)<\cdots<U_{n-1}(r_{n-1})$. Every local $q$-particle Softmin saddle ($2\le q\le n-1$) has strictly larger action than the corresponding one-particle saddle, and hence larger action than $V_1$.
    
    It remains to exclude the fully synchronous channel $q=n$. If all particles leave $D$ through the boundary bottleneck, then by minimality of $x^\ast$ on $\partial D$, $f(x_i)\ge f^\ast$ for all $i$, yielding $nJ_\beta(\mathbf{x})\ge n f^\ast$. For $n\ge2$, we have 
    \[
        \overline{V}_\beta \le B_\beta := \max\left\{ \frac1\beta, \frac{\beta}{2}\left(f^\ast-\frac1\beta\right)_+^2 \right\}.
    \]
    Choose $n_0$ so that $n f^\ast>B_\beta$ for all $n\ge n_0$. Then, for all such $n$, $n f^\ast > B_\beta \ge \overline{V}_\beta \ge V_1$. Thus the fully synchronous channel is also strictly more expensive, completing the lemma.
\end{proof}

\begin{proof}[Proof of Theorem \ref{thm:transition-time}]
\label{thm:transition-time-proof}
    The process \eqref{eq:gradient-flow-noisy} is a small-noise gradient diffusion in $\mathbb{R}^{nd}$ with potential $nJ_\beta$. Its Freidlin-Wentzell rate functional is $I_{0T}$. Therefore, the logarithmic scale of an exit from the metastable well is the minimum action needed to reach the exit set.
    
    By Lemma \ref{lem:single-particle-channel}, for all $n\ge n_0$, $V_q > V_1$ for $q=2,\ldots,n$. Set $V_{\ge2}:=\min_{2\le q\le n} V_q$. Since $V_{\ge2}>V_1$, there exists $\gamma>0$ such that $V_{\ge2}\ge V_1+\gamma$.
    
    The Freidlin-Wentzell upper and lower bounds imply that, for every sufficiently small $\eta>0$, the probability of a multi-particle transition is bounded by $\mathbb{P}_{\bar x}(\text{multi-particle}) \lesssim \exp\left(-(V_1+\gamma-\eta)/\sigma\right)$, whereas the one-particle transition occurs on the scale $\exp\left(-(V_1+\eta)/\sigma\right)$. Taking the ratio:
    \[
        \frac{ \mathbb{P}_{\bar x}(\text{multi-particle transition}) }{ \mathbb{P}_{\bar x}(\text{single-particle transition}) } \lesssim \exp\left(-\frac{\gamma-2\eta}{\sigma}\right) \longrightarrow 0 \quad \text{as } \sigma \downarrow 0.
    \]
    Consequently, exactly one particle exits during the transition with probability tending to one.
    
    It remains to justify that the other particles stay localized during the actual transition. Since $x_{\min}$ is a nondegenerate local minimum, the standard Freidlin-Wentzell confinement estimate dictates that for every $\delta>0$, there exist constants $C_\delta,c_\delta>0$ such that, for every transition window $T_\sigma=\mathcal{O}(|\log\sigma|)$,
    \[
        \mathbb{P}\left( \sup_{0\le t\le T_\sigma} \|X_t^{j,\sigma}-x_{\min}\|>\delta \right) \le C_\delta T_\sigma e^{-c_\delta/\sigma}.
    \]
    Since $T_\sigma e^{-c_\delta/\sigma}\to0$, a union bound over the nonescaping particles shows that all of them remain in an arbitrarily small neighborhood of $x_{\min}$ during the short transition window with probability tending to one.
    
    The Freidlin-Wentzell exit-time theorem then gives $\lim_{\sigma\downarrow0} \sigma\log \mathbb{E}_{\bar x}\tau_{\mathrm{any}}^\sigma = V_1$. By exchangeability, the fixed tagged-particle exit time differs from the first-particle exit time only at the prefactor level. Therefore, for every fixed $k$, $\lim_{\sigma\downarrow0} \sigma\log \mathbb{E}_{\bar x}\tau_k^\sigma = V_1$.
    
    Finally, Lemma \ref{lem:single-particle-channel} establishes $V_1\le\overline{V}_\beta$. Hence, for every $\eta>0$ and all sufficiently small $\sigma$,
    \[
        \mathbb{E}_{\bar x}\tau_{\mathrm{any}}^\sigma \le \exp\left(\frac{\overline{V}_\beta+\eta}{\sigma}\right), \qquad \mathbb{E}_{\bar x}\tau_k^\sigma \le \exp\left(\frac{\overline{V}_\beta+\eta}{\sigma}\right).
    \]
    This completes the proof.
\end{proof}

\color{black}

\begin{lemma}
    \label{lem:minimizing-particle-proof}
    If a particle $x_k$ is minimizing (resp., strongly minimizing) at time $t=0$, then under the deterministic gradient flow dynamics (Eq. \ref{eq:gradient-flow}), it remains minimizing (resp., strongly minimizing) for all $t>0$.
\end{lemma}
\begin{proof}
    Consider the time derivative of the effective learning rate $\prefactor{t}{k}$. Assuming the non-exploring ensemble remains locally stationary (as justified in Remark 1), the temporal evolution of $\prefactor{t}{k}$ is governed by the chain rule with respect to the $k$-th particle's objective value:
    \begin{equation*}
        \frac{d}{dt}\prefactor{t}{k} = \frac{\partial \prefactor{t}{k}}{\partial f(x_k)} \frac{d}{dt}f(x_k)
    \end{equation*}
    Under the deterministic Softmin dynamics $\dot{x}_k = -\prefactor{t}{k} \nabla f(x_k)$, the instantaneous change in the objective function is:
    \[
        \frac{d}{dt} f(x_k) = \langle \nabla f(x_k), \dot{x}_k \rangle = -\prefactor{t}{k} \|\nabla f(x_k)\|^2
    \]
    If the particle is in the minimizing regime at time $t$ ($\prefactor{t}{k}>0$), the objective value strictly decreases (or is stationary at a critical point), yielding $\frac{d}{dt} f(x_k) \leq 0$. Furthermore, based on the formulation of the effective learning rate, the partial derivative with respect to the particle's objective value is strictly negative ($\frac{\partial}{\partial f(x_k)}\prefactor{t}{k} < 0$).

    Consequently, the product of these two non-positive terms implies that the time derivative is non-negative:
    \[
        \frac{d}{dt}\prefactor{t}{k} \geq 0
    \]
    This establishes that $\prefactor{t}{k}$ is monotonically non-decreasing over time. Since the particle is initially minimizing ($\prefactor{0}{k} > 0$), it must hold that $\prefactor{t}{k} > 0$ for all $t > 0$. 
    
    By an identical argument, if the particle is initially strongly minimizing ($\prefactor{0}{k} \geq 1$), this monotonic non-decrease guarantees that $\prefactor{t}{k} \geq 1$ for all $t > 0$, successfully trapping the particle in the strongly minimizing regime.
\end{proof}

\begin{theorem}
    \label{thm:rate-of-convergence-minimizing-particle-proof}
    Let $f$ be locally strongly convex and $x_k(0)$ be a strongly minimizing particle, so that $\prefactor{0}{k}>1$. Assume that $x_k(0),x^*\in K$ where $x^*\in \text{argmin }f$ and $K$ is compact. Then, the deterministic dynamics in \eqref{eq:gradient-flow} converges to $x^*$ at a rate $\mathcal{O}(e^{-\lambda t})$, where $\lambda$ is the strong convexity constant of the function $f$ in the compact set $K$.
\end{theorem}
\begin{proof}
    From Lemma \ref{lem:minimizing-particle}, we know that $x_k(t)$ remains strongly minimizing throughout the entire trajectory. Tracking the squared distance to the minimum, we have:
    \begin{align*}
        \frac{1}{2}\frac{d}{dt}\left\|x_k(t)-x^*\right\|^2 & = \left\langle x_k(t)-x^*, \dot{x}_k(t) \right\rangle \\
        & = \left\langle x_k(t)-x^*, -\prefactor{t}{k}\nabla f(x_k(t)) \right\rangle.
    \end{align*}
    From the definition of local strong convexity (Definition \ref{def:local-strong-convexity}) and the fact that $x^*$ is a local minimum ($\nabla f(x^*) = 0$), we obtain the standard lower bound on the inner product:
    \[
        \left\langle x_k(t)-x^*, \nabla f(x_k(t)) \right\rangle \geq f(x_k(t)) - f(x^*) + \frac{\lambda}{2} \| x_k(t)-x^* \|^2 \geq 0.
    \]
    Because the particle is strongly minimizing, we established that $\prefactor{t}{k} > 1$. Coupling this with the positivity of the inner product, we can bound the time derivative as follows:
    \begin{align*}
        \frac{1}{2}\frac{d}{dt}\left\|x_k(t)-x^*\right\|^2 & = -\prefactor{t}{k} \left\langle x_k(t)-x^*, \nabla f(x_k(t)) \right\rangle \\
        & \leq -\left\langle x_k(t)-x^*, \nabla f(x_k(t)) \right\rangle \\
        & = -\left\langle x_k(t)-x^*, \nabla f(x_k(t)) - \nabla f(x^*) \right\rangle \\
        & \leq -\lambda\left\|x_k(t)-x^*\right\|^2,
    \end{align*}
    where the final step follows directly from the monotonicity of the gradient for $\lambda$-strongly convex functions. 
    
    This yields the differential inequality $\frac{d}{dt}\|x_k(t)-x^*\|^2 \leq -2\lambda\|x_k(t)-x^*\|^2$. Applying Grönwall's inequality immediately yields the exponential convergence rate:
    \[
        \left\|x_k(t)-x^*\right\|^2 \leq \left\|x_k(0)-x^*\right\|^2 e^{-2\lambda t}. \qedhere
    \]
\end{proof}

\section{Appendix: Experimental Details and Hyperparameters}
\label{app:hyperparameters}

To ensure full reproducibility of the numerical experiments presented in Section \ref{sec:numerical-experiments}, this appendix details the hyperparameters and initialization schemes used across all benchmark landscapes. All continuous-time dynamics were discretized using the Euler-Maruyama method. The source code used to execute these experiments is publicly available at \cite{softmin_energy_2026}.

\subsection{Hardware and Computational Resources}
All numerical experiments were executed on the \emph{Toeplitz} Scientific Computing Cluster hosted by the Department of Mathematics at the University of Pisa. To fully leverage the vectorized nature of the simulations and evaluate independent trials in parallel, jobs were strictly allocated to a single compute node at a time. The cluster nodes utilized for these experiments include:
\begin{itemize}
    \item \textbf{High-Capacity Nodes:} AMD EPYC 7763 processors featuring 128 physical cores (64 cores per socket, 2 sockets, 256 logical threads) and 2 TB of RAM. 
    \item \textbf{Standard Nodes:} Intel Xeon CPU E5-2650 v4 @ 2.20GHz featuring 24 physical cores (12 cores per socket, 2 sockets) and 256 GB of RAM. These nodes were specifically utilized to parallelize the independent runs of the experiments using Python's \texttt{multiprocessing} framework.
\end{itemize}

\subsection{General Experimental Settings}
Table \ref{tab:general_settings} summarizes the shared parameters utilized across all algorithms for each specific landscape. The swarm size and maximum step limits were kept constant to provide a fair baseline for evaluating the topological escape mechanisms.

\begin{table}[!htbp]
\centering
\caption{General settings and landscape configurations.}
\label{tab:general_settings}
\begin{tabular}{@{}ll@{}}
\toprule
\textbf{Parameter} & \textbf{Value / Description} \\ 
\midrule
Swarm Size ($n$) & 100 \\
Independent Runs & 100 \\
Max Step Limit & 10,000,000 \\
\midrule
\textbf{1D Double Well} & $f(x) = \frac{1}{4}x^4 - a x^2$ \\
Initialization & Uniformly sampled around the left-most minimum \\
Domain Bounds & $[-\infty, \infty]$ \\
\midrule
\textbf{2D Quadruple Well} & $f(x,y) = x^4 + y^4 - a(x^2 + y^2) - xy$ \\
Initialization & Uniformly sampled around the suboptimal trap \\
Domain Bounds & $[-4.0, 4.0]^2$ \\
\midrule
\textbf{High-Dimensional Ackley} & Generalized Ackley ($d \in [10, 90]$) \\
Function Parameters & $a=20.0$, $b=0.2$, $c=2\pi$ \\
Initialization & Uniformly sampled in $[-10, 10]^d$ \\
Domain Bounds & $[-10, 10]^d$ \\
\bottomrule
\end{tabular}
\end{table}

\subsection{Algorithm-Specific Hyperparameters}

The following tables detail the algorithm-specific hyperparameters. For methods utilizing a decaying schedule (Softmin Decaying, Simulated Annealing), the relevant parameter was updated at each step $t$ according to a geometric decay schedule of the form $P_t = P_0 \times (\text{decay rate})^t$.

\begin{table}[!htbp]
\centering
\caption{Hyperparameters for Softmin Energy Dynamics.}
\label{tab:softmin_hyperparams}
\begin{tabular}{@{}ll@{}}
\toprule
\textbf{Parameter} & \textbf{Value} \\ 
\midrule
Learning Rate ($\eta$) & 0.01 \\
Adaptive Learning Rate (Adagrad) & Disabled (False) \\
Noise Intensity ($\sigma$) & 1.0 (Benchmarks), 0.05 (Theorem 3.1 Verification) \\
Noise Distribution & Normal \\
Fixed Inverse Temperature ($\beta$) & 10.0 \\
Decaying Inverse Temperature ($\beta_0$) & 10.0 \\
Geometric Decay Rate for $\beta$ & 0.9999 \\
\bottomrule
\end{tabular}
\end{table}

\begin{table}[!htbp]
\centering
\caption{Hyperparameters for Simulated Annealing (SA).}
\label{tab:sa_hyperparams}
\begin{tabular}{@{}ll@{}}
\toprule
\textbf{Parameter} & \textbf{Value} \\ 
\midrule
Learning Rate ($\eta$) & 0.01 \\
Base Noise Variance Factor & $2 \times \sigma \times \eta$ \\
Initial Temperature ($T_0$) & 10.0 \\
Cooling Schedule & Geometric \\
Geometric Cooling Rate & 0.9999 \\
\bottomrule
\end{tabular}
\end{table}

\begin{table}[!htbp]
\centering
\caption{Hyperparameters for Gradient-Aware Particle Swarm Optimization (PSO).}
\label{tab:pso_hyperparams}
\begin{tabular}{@{}ll@{}}
\toprule
\textbf{Parameter} & \textbf{Value} \\ 
\midrule
Gradient Step Learning Rate ($\eta$) & 0.01 \\
Inertia Weight ($w$) & 0.729 \\
Cognitive Coefficient ($c_1$) & 1.49445 \\
Social Coefficient ($c_2$) & 1.49445 \\
Injected Noise Floor & 0.5 \\
\bottomrule
\end{tabular}
\end{table}

\begin{table}[!htbp]
\centering
\caption{Hyperparameters for Consensus-Based Optimization (CBO).}
\label{tab:cbo_hyperparams}
\begin{tabular}{@{}ll@{}}
\toprule
\textbf{Parameter} & \textbf{Value} \\ 
\midrule
Learning Rate ($\Delta t$) & 0.01 \\
Consensus Weight ($\lambda$) & 1.0 \\
Exploration Noise ($\sigma_{\text{CBO}}$) & $\sqrt{2}$ \\
Temperature ($\alpha$) & 10.0 \\
\bottomrule
\end{tabular}
\end{table}

\subsection{Further Experimental Results}
\label{apx:other-experiments}

\begin{figure*}
    \centering
    \includegraphics[width=0.8\linewidth]{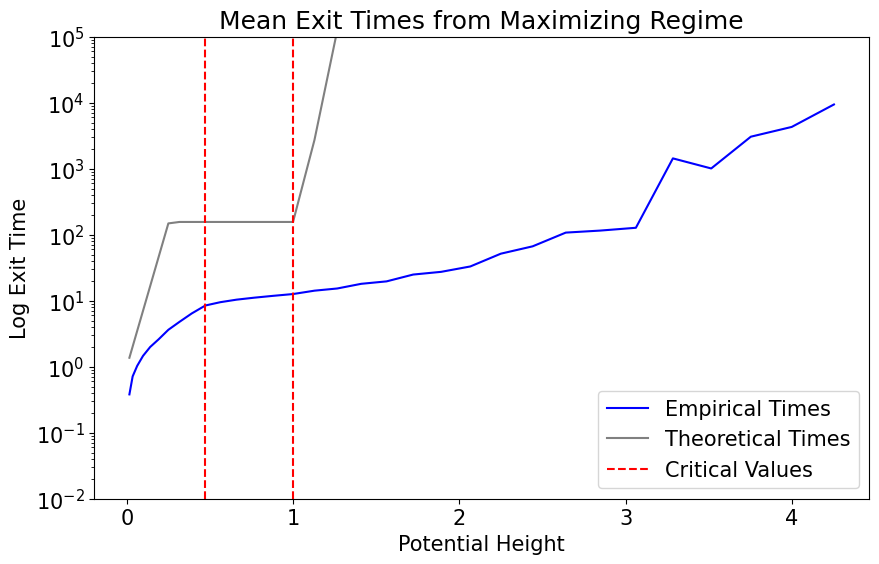}
    \caption{Empirical exit times strictly follow the bound from Theorem \ref{thm:transition-time}}
    \label{fig:theory}
\end{figure*}

Figure \ref{fig:theory} compares the empirical exit times against the theoretical upper bound $\overline{V}_\beta$. Dashed lines demarcate phase transitions where the dominant term governing the exit time shifts: from $f^*$ to $n/(2(n-1)\beta)$, and ultimately to $(\beta/2)(f^* - 1/\beta)^2$. Validating our theoretical predictions, the empirical slope changes exactly at these boundaries, most prominently at the first transition. Furthermore, while the empirical times remain bounded by $\overline{V}_\beta$ across all regimes, the theoretical bound is not tight in the third phase.


\end{document}